\documentclass[letterpaper, 10 pt, conference]{ieeeconf}  

\IEEEoverridecommandlockouts

\usepackage{cite}

\usepackage{graphics} 
\usepackage{epsfig} 
\usepackage{mathptmx} 
\usepackage{times} 
\usepackage{amsmath} 
\usepackage{amssymb, mathtools}  

\makeatletter
\def\@opargbegintheorem#1#2#3{\trivlist
   \item[]{\bfseries #1\ #2\ (#3)} \itshape}
\makeatother
\newtheorem{theorem}{\bf Theorem}[section]
\newtheorem{definition}[theorem]{\bf Definition}
\DeclareMathOperator*{\argmin}{arg\,min}

\usepackage{float}
\usepackage[breaklinks=true,bookmarks=true,colorlinks]{hyperref}
\usepackage{graphicx}
\usepackage{xcolor}
\usepackage{multirow}  
\usepackage{adjustbox}
\usepackage{float}
\usepackage{subcaption}

\title{\LARGE \bf
RNBF: Real-Time RGB-D Based Neural Barrier Functions\\ for Safe Robotic Navigation}

\author{Satyajeet Das$^{*1}$, Yifan Xue$^{2}$, Haoming Li$^{2}$  and Nadia Figueroa$^{2}$
\thanks{*Corresponding author.}
\thanks{$^{*1}$Satyajeet Das is with the University of Southern California, Los Angeles, CA, 
90007 USA, {\tt\scriptsize\ satyajee@usc.edu}. This research was primarily conducted while at the University of Pennsylvania.}%
\thanks{$^{2}$Yifan Xue, Haoming Li, and Nadia Figueroa are with the University of Pennsylvania, Philadelphia, PA,
19104 USA, {\tt\scriptsize\ yifanxue@seas.upenn.edu, lihaomingforreal@gmail.com, nadiafig@seas.upenn.edu}.}%
}

\begin{document}

\maketitle
\thispagestyle{empty}
\pagestyle{empty}

\begin{abstract}

%

Autonomous safe navigation in unstructured and novel environments poses significant challenges, especially when environment information can only be provided through low-cost vision sensors. Although safe reactive approaches have been proposed to ensure robot safety in complex environments, many base their theory off the assumption that the robot has prior knowledge on obstacle locations and geometries.
In this paper, we present a real-time, vision-based framework that constructs continuous, first-order differentiable Signed Distance Fields (SDFs) of unknown environments entirely online, without any pre-training, and is fully compatible with established SDF-based reactive controllers.
To achieve robust performance under practical sensing conditions, our approach explicitly accounts for noise in affordable RGB-D cameras, refining the neural SDF representation online for smoother geometry and stable gradient estimates. 
We validate the proposed method in simulation and real-world experiments using a Fetch robot. Videos and supplementary material are available at \url{https://satyajeetburla.github.io/rnbf/}.
 \end{abstract}
\section{INTRODUCTION}
Safety in robotics is a critical challenge, especially when operating in unstructured environments where obstacle states and geometries are unavailable \cite{yang2018grand}. In recent years, Control Barrier Functions (CBFs) were widely studied due to their ability to provide safety guarantees for any control affine systems and adopted in various applications including vehicle cruise control \cite{cbfcruise}, indoor navigation \cite{xue2025minimacollisionscombiningmodulation}, quadrotor control \cite{cbfquadrotor}, legged robot locomotion \cite{grandia2021multi}, etc. 
All the aforementioned works assume that the robot has prior knowledge of obstacle states and geometries, or even a predefined map. However, for robot navigation in unknown environments, such assumptions are unrealistic and the robot can only access noisy environment information from its onboard sensors. 

CBFs's performance in enforcing robot safety relies not only on the robot's ability to collect accurate nearby obstacle state information but also on how well the collected information is processed to meet the requirement of valid CBFs. Therefore, when environment descriptions are unavailable or noisy, the safety guarantee of robot navigation is jeopardized. Building valid CBFs adaptively using noisy data collected from robots' on-board sensors in real time remains a challenge.  
Many recent works use LiDAR sensors to construct CBFs and ensure robot safety in unknown environments \cite{zhang2024online, dawson2022learning, long2021learning, harms2024neural}. 
In this work, we instead demonstrate that safe reactive control can be achieved with RGB-D camera alone, without relying on additional sensing modalities such as LiDAR.
\begin{figure}
    \centering
    \includegraphics[width=\linewidth]{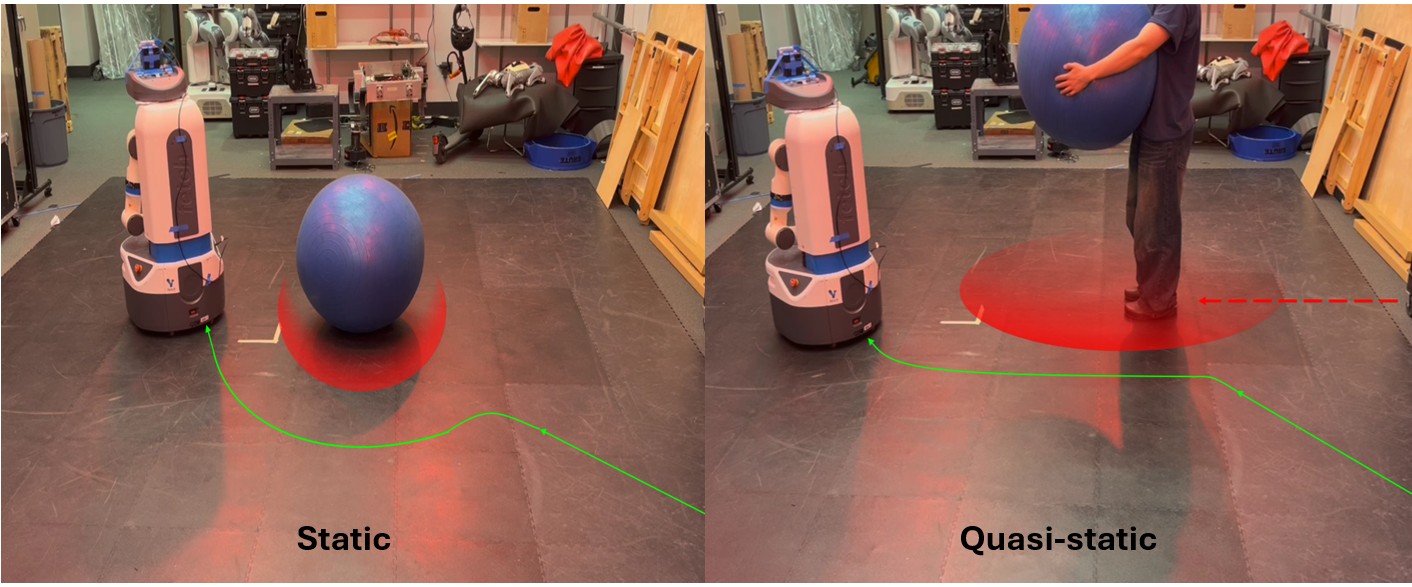}
    
     \caption{Hardware setup for real-world experiments showing the RNBF-Control pipeline navigating around static and quasi-static obstacles without any prior knowledge about the obstacles.}
    \label{fig:fetch}
    \vspace{-5mm}
\end{figure}

Seldom works have investigated perception-based CBFs using RGB-D camera inputs. Existing vision-based CBF pipelines utilize machine-learning techniques to incorporate robot observations into CBF formulations. For example, \cite{abdi2023safe} leverages a conditional Generative Adversarial Network to approximate CBFs given RGB-D camera inputs; \cite{tong2023enforcing} predicts future RGB-D observations with Neural Radiance Fields (NeRFs) and feeds them into discrete-time CBFs.
These methods require significant offline training and their performance can be affected by the quality of their training dataset. 
Additionally, using CBFs in discrete-time settings may weaken their safety guarantees when the update frequency is not fast enough. Other works, like \cite{de2024point}, approximate mathematically CBFs and their gradients from point cloud data collected through LiDARs or depth cameras. 

Among the methods listed above, it is noteworthy that none has explicitly considered the effects of sensory noises on perception-based CBF. Further, most existing vision-based CBFs have made approximations to standard CBFs to accommodate robot observation formats. Assumptions that CBFs are founded on, such as the continuity and first-order differentiability of the barrier function are typically relaxed, weakening their safety guarantees. Additionally, altering standard CBFs in vision-based safe navigation hinders the adaptation of well-developed none-sensor-based safe reactive control algorithms into vision-based robot platforms. Many safe reactive controllers, including Control Barrier Function Quadratic Programs (CBF-QP), are developed assuming the system's access to SDFs and their gradients of the obstacles near the robot \cite{zucker2013chomp},\cite{cbfcruise},\cite{cbfstar},\cite{LukesDS}, \cite{onManifoldMod} and \cite{xue2025minimacollisionscombiningmodulation}. In robot safe navigation, SDFs refer to the orthogonal distance of a given point to the boundary of an unsafe set. 

Therefore, in this paper, we offer an online barrier function generation framework capable of smoothing sensor noises and constructing continuous and first-order differentiable neural SDFs in real-time given RGB-D camera inputs without pre-training for any unknown environments. 
With the advancements presented by NeRF\cite{mildenhall2021nerf},  the quality and efficiency of environmental reconstruction have significantly improved. However, limited work explored generating SDFs using NeRF, most of which were designed for an offline setting \cite{liu2022regularized,tang2023rgb}.  A notable exception is iSDF\cite{ortiz2022isdf}, which introduced a method for real-time SDF generation.  
Recently, \cite{bukhari2025differentiable} proposed a dual-level SDF approach combining pretrained object SDFs with a scene-level iSDF for indoor navigation. Their method focused on trajectory optimization and relies on offline components, without incorporating any safety-filtering framework.
In contrast, our vision pipeline generates real-time, continuous, and differentiable SDFs, specifically designed for constructing continuous CBFs for safe robot navigation.
Instead of altering CBFs to accommodate vision observations, as in prior works, we establish continuous and first-order differentiable SDFs online, which enables our methods to be readily adaptable to SDF-based safe reactive controllers such as standard CBF-QPs \cite{ames2019control, xue2025no} and dynamical system based modulation \cite{fourie2024manifold,koptev2024reactive}.
Our approach is validated in both gazebo simulations and real-life hardware experiments using the Fetch robot in static and quasi-static obstacle environments. 

Our \textbf{key contributions} are as follows:
\begin{itemize}
\item We introduce an approach to generate real-time (5-15 Hz) SDFs from raw RGB-D data using an online neural representation compatible with established SDF-based reactive controllers.
\item To account for depth noise inherent in affordable sensors like RGB-D cameras, we mitigate its impact on neural SDF representations adopting the huber loss~\cite{huber1992robust}.
\item Leveraging the neural SDF representation, we propose a method to construct barrier functions in real-time without prior training, leveraging cost-effective RGB-D camera for safe navigation in novel environments.
\end{itemize}

\section{Problem Formulation}
Consider a control-affine nonlinear system of the form
\begin{equation}
    \dot{x} = f(x) + g(x)u \label{eq:affine system},
\end{equation}
where $x \in \mathbb{R}^n$ is a state vector, $u \in \mathbb{R}^m$ is a control input vector, $f:\mathbb{R}^n \to \mathbb{R}^n$, and $g:\mathbb{R}^n\to \mathbb{R}^{n \times m}$ define the control-affine nonlinear dynamics. We define the notion of robot safety and consider safe control algorithms based on barrier functions. 

Given a continuously differentiable function $h:\mathbb{R}^n \rightarrow \mathbb{R}$, $h$ is a barrier function if the safe set $C$ (outside the obstacle), the boundary set $\partial C$ (on the boundary of the obstacle), and the unsafe set $\neg C$ (inside the obstacles) of the system can be defined as in \eqref{eq:safe region o},~\eqref{eq:boundary o},~\eqref{eq:unsafe region o}~\cite{ames2019control}. 
\begin{equation}
C=\{x \in \mathbb{R}^n: h(x)>0\}\label{eq:safe region o}
\end{equation}
\begin{equation}
\partial C=\{x \in \mathbb{R}^n: h(x)=0\}\label{eq:boundary o}
\end{equation}
\begin{equation}
\neg C=\{x \in \mathbb{R}^n: h(x)<0\}\label{eq:unsafe region o}
\end{equation}

In practice, $h(x)$ is often measured as the shortest distance from state $x$ to the boundary set $\partial C$, i.e. a signed distance function (SDF). 

\textbf{Assumptions:} We assume the robot has access to its poses (through odometry or external sensing) in real-time during the vision-based navigation. The entire vision-based navigation pipeline takes in 2 poses: the camera pose $x_\text{cam}$ and the robot actuation center pose $x_\text{rob}$ measured in the world frame. Given $h(x)$ that returns the Euclidean distance to the nearest obstacle, it can be deduced that $|h(x_\text{rob})-h(x_\text{cam})|<\zeta$, where $\zeta$ is the Euclidean distance between the robot and camera centers. Since our framework is designed specifically for robots with onboard RGB-D cameras, where $\zeta$ is a small restricted value, we assume that $h(x_\text{rob}) \approx h(x_\text{cam}) = h(x)$ and ensure robot safety by defining a safety margin greater or equal to $\zeta$. 

\textbf{Problem Statement:} Given a robot with accurate knowledge of its onboard camera poses in an unexplored environment, the goal of our vision-based navigation pipeline is to (i) take in noisy RGB-D camera inputs, (ii) generate continuous and first-order differentiable SDFs as barrier functions and (iii) compute an admissible input $u$ that will ensure the state of the robot $x$ is always within the safe set $C$ defined in \eqref{eq:safe region o} using standard CBF-QP.

\label{sec: assumptions}



\begin{figure*}[tbp]
\centering
\includegraphics[width=\textwidth]{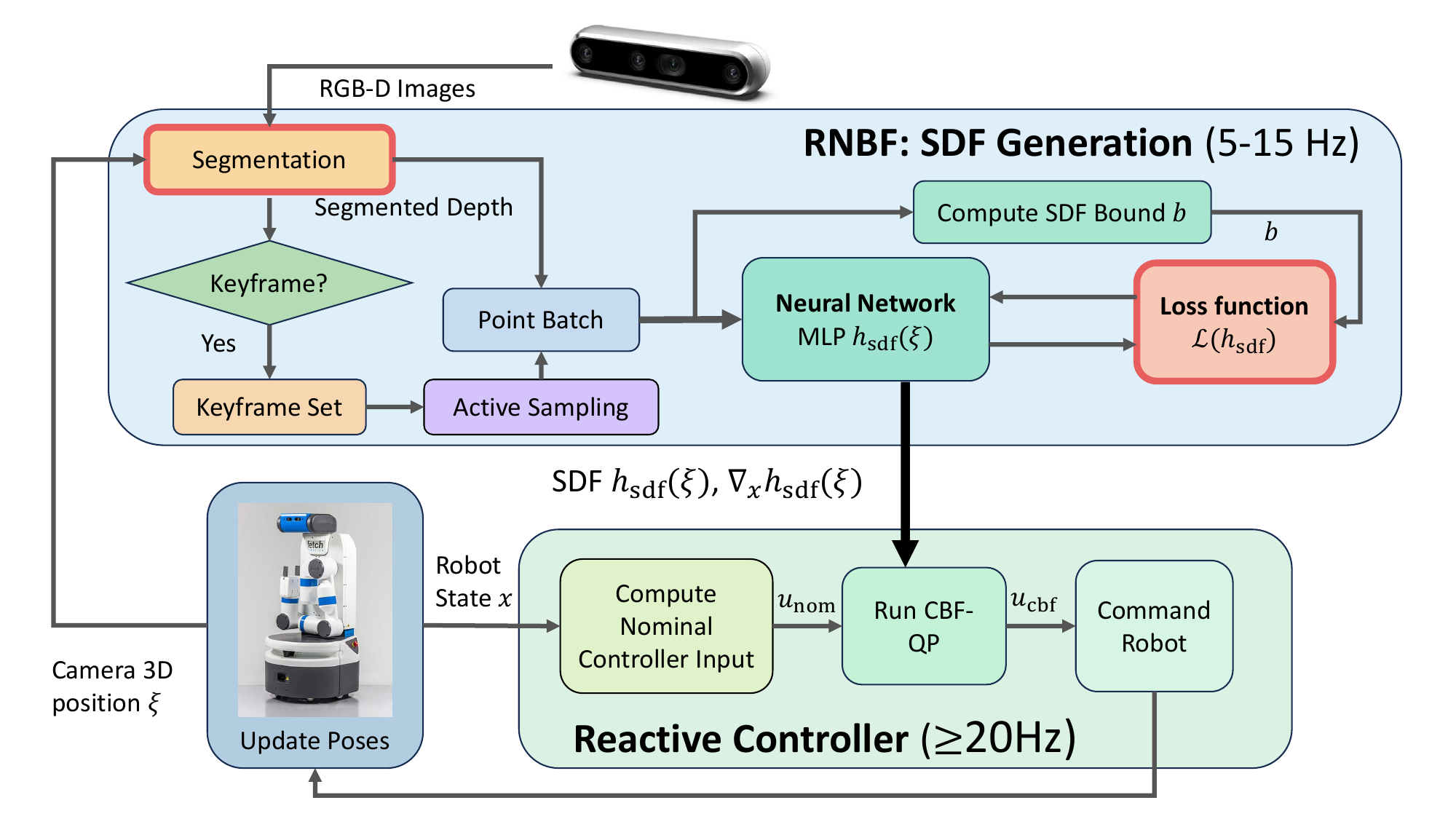}
\caption{
Block diagram of the proposed RNBF-based safe reactive navigation pipeline. The SDF Generation module (5–15 Hz) constructs a continuous, differentiable signed distance field from real-time RGB-D input, while the Reactive Controller ($\geq 20$\,Hz) ensures collision-free navigation using a CBF-QP. The black thick arrow highlights the main coupling step between the perception and control pipelines. Red outlines highlight components in the SDF generation stage that distinguish the RNBF pipeline from the original iSDF pipeline.}
\label{fig:sdf_pipelinef}
\vspace{-10pt}
\end{figure*}
\section{PRELIMINARY}
\subsection{Control Barrier Function}
\label{sec: CBF}
 To guarantee safety of the controlled agent, CBF-QP utilizes the Nagumo Set Invariance Theorem.

\begin{definition}[Nagumo Set Invariance]
\begin{equation}
\label{Nagumo}
C \text{ is set invariant} \iff \Dot{h}(x) \geq 0 \; \forall x \in \partial C\\
\end{equation}
\end{definition}

Control Barrier Functions are designed by extending Nagumo set invariance theorem to a ``control" version\cite{ames2019control}, where the condition $\forall x \in \partial C$ is rewritten mathematically using an extended $\mathcal{K}_\infty$ function $\alpha$, 
\begin{equation}
\label{eq:CBF conditions}
 C \text{ is set invariant} \iff \exists u \; \text{s.t.} \; \Dot{h}(x, u) \geq  -\alpha (h(x)).
\end{equation}

\begin{definition}[Extended $K_{\infty}$ Functions]
\label{def:K inf}
An extended $\mathcal{K}_{\infty}$ function is a function $\alpha : \mathbb{R}\rightarrow \mathbb{R}$ that is strictly increasing and with $\alpha (0)=0$ ; that is, extended $\mathcal{K}_{\infty}$ functions are defined on the entire real line $(-\infty, \infty) $.
\end{definition}
The CBF condition in \eqref{eq:CBF conditions} can be used to formulate a quadratic programming (QP) problem that guarantees safety by enforcing the set invariance of the safety set $C$ defined in \eqref{eq:safe region o}. For general control affine systems as in \eqref{eq:affine system}, CBF-QP is defined as
\begin{gather}
\label{eq:cbf-qp affine}
u_{\text{cbf}} = \argmin_{{u} \in \mathbb{R}^m}\frac{1}{2}||u-u_{\text{nom}}||_2^2\\
\label{eq:cbf-qp affine_1}
L_fh(x) + L_gh(x)u \geq -\alpha (h(x))
\end{gather}

\subsection{Signed Distance Function (SDF)}
When representing three-dimensional geometry, a signed distance field $h_\text{sdf}:\mathbb{R}^3 \rightarrow \mathbb{R}$ is a scalar field that maps a 3D coordinate $\xi = [p_x, p_y, p_z]^\top$ to the scalar signed distance value \cite{ortiz2022isdf}.  The surface $S$ is the zero
level-set of the field:
\begin{equation}\label{sdf_equa}
S = \{\xi \in \mathbb{R}^3 \mid h_\text{sdf}(\xi) = 0\}
\end{equation}

By definition, signed distance fields (SDF) satisfy the requirement of barrier functions in Eq.~\eqref{eq:safe region o},\eqref{eq:boundary o} and \eqref{eq:unsafe region o} when $n=3$. Signed distance fields are good barrier function candidates because they retain differentiability almost everywhere (depending on the formulation), and their gradients are identical to their surface normals. 

\subsection{iSDF: Real-time SDF Reconstruction}
\label{sec:prelim-isdf}

iSDF\cite{ortiz2022isdf} proposed a continual learning system for real-time estimates a SDF from a stream of depth images with known camera poses.  
The SDF is represented by a multilayer perceptron (MLP) mapping a 3D location to its signed distance.
At each iteration, the system selects a small set of depth frames via active keyframe replay to mitigate catastrophic forgetting. For each selected frame, pixels and depths are sampled along back-projected rays to form a batch of 3D query points. The model is evaluated at these points and trained with a self-supervised objective; parameters are updated by backpropagation. The work reported real-time performance and evaluates the method on real and synthetic indoor datasets\cite{ortiz2022isdf}.

\vspace{2pt}

\section{Methodology}

\subsection{SDF Generation Pipeline}
\label{subsec:SDF_Generation} 

\subsubsection{Core Methodology}

Our real-time SDF pipeline (\autoref{fig:sdf_pipelinef}, “SDF Generation”) follows the iSDF\cite{ortiz2022isdf} template but differs primarily in two major components that are critical for safe robot navigation: 
(i) an explicit \emph{floor segmentation} stage 
(ii) a \emph{robust near–surface loss} (Sec.~\ref{subsubsec:loss}) that replaces the $L_1$ term used in iSDF to attenuate RGB-D depth noise.

Given a sequence of posed RGB-D images, Our pipeline first segments the floor and masks the corresponding depth pixels to prevent the large planar ground from being treated as an obstacle, following standard RGB-D scene understanding practice~\cite{gupta2013perceptual}. The segmentation module is modular, and any modern method (e.g., SAM~\cite{kirillov2023segment}) can be used. In our implementation, for simplicity we use a fast color-based segmenter on the RGB image and apply the resulting mask to the depth image. At the heart of our pipeline also lies a MLP, which transforms a three-dimensional point $\xi$ into its corresponding signed distance. The network (Sec.~\ref{subsubsec:neural_arch}) is randomly initialized and refined online from the incoming masked depth stream. We also maintain a small keyframe set, select a few frames per update via active replay, and sample query points along back-projected rays using the sampling strategy as in iSDF\cite{ortiz2022isdf}. The neural SDF $h_{\text{sdf}}(\xi)$ is queried at these sampled points and optimized with the objective in section~\ref{subsubsec:loss}, yielding a continuous SDF and gradients available online (5--15\,Hz) to the reactive safety controller (running at $\geq 20$\,Hz) for safe robot navigation.

\vspace{2pt}

\begin{figure}[h!]
\includegraphics[width=\linewidth]{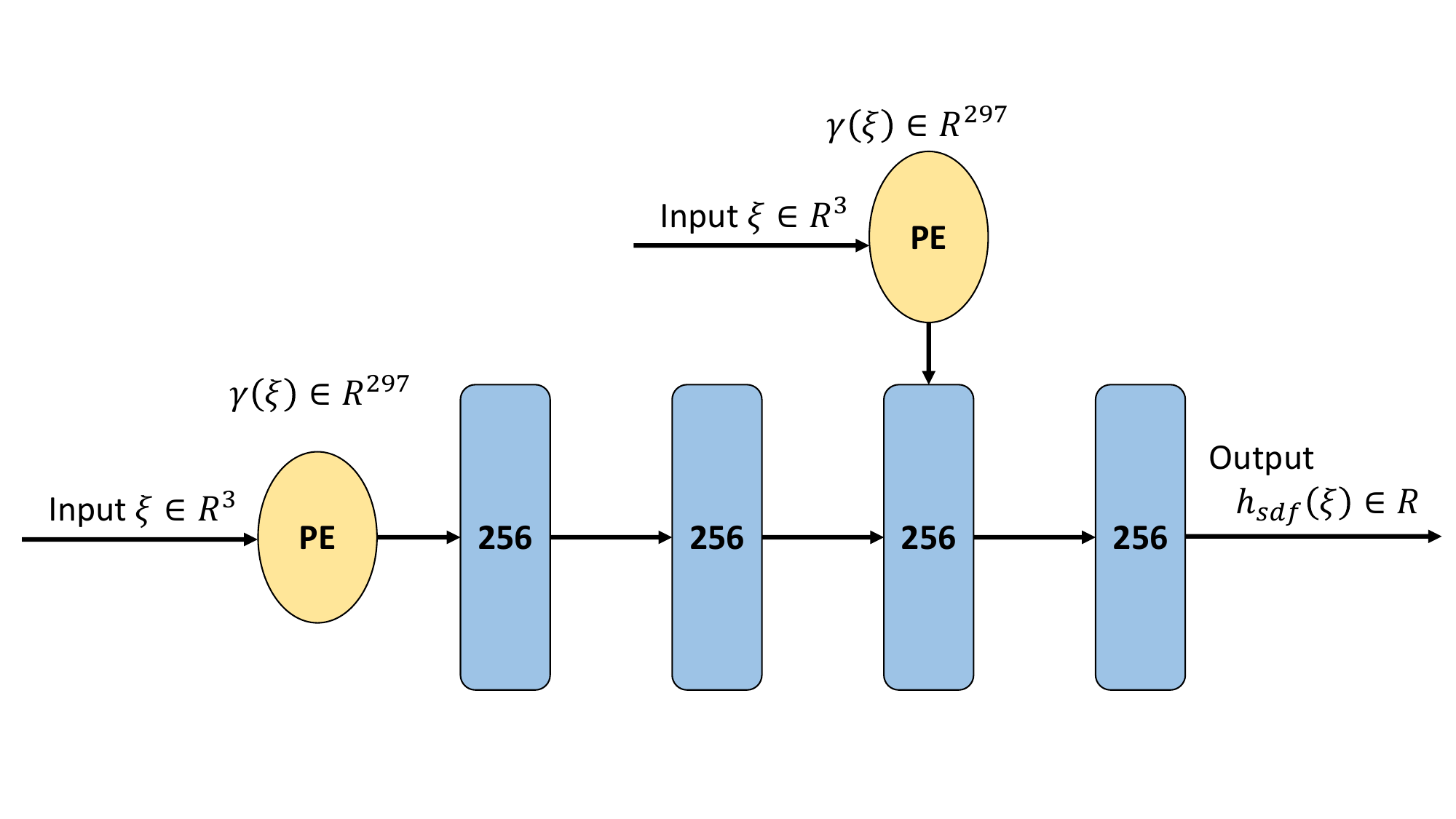}
\caption{Architecture of the neural SDF network \(h_\text{sdf}(\xi)\). The model is a fully connected MLP composed of four hidden layers, each with 256 units (depicted as blue rectangles). Yellow elliptical blocks marked “PE” represent positional encodings, where input coordinates \(\xi \in \mathbb{R}^3\) are transformed via a periodic embedding \(\gamma(\xi) \in \mathbb{R}^{297}\) to enable high-frequency detail reconstruction. The final linear layer outputs the scalar signed distance value \(h_\text{sdf}(\xi) \in \mathbb{R}\).}

\label{fig:MLP_sdf}
\end{figure}


\subsubsection{Neural Architecture}
\label{subsubsec:neural_arch} 
The neural SDF (\autoref{fig:MLP_sdf})  is defined by a parameterized MLP, \( h_\text{sdf}(\xi;\theta) \), where
\(\xi \in \mathbb{R}^3\) denotes the input coordinates, \(H\) denotes the hidden layers of MLP and 
\(\theta = \{\mathbf{W}^{(i)}, \mathbf{c}^{(i)}\}_{i=1}^H \cup \{\mathbf{W}, \mathbf{c}\}\) 
represents the learnable weight matrices and biases. For brevity, we will write \(h_\text{sdf}(\xi)\) instead of \(h_\text{sdf}(\xi;\theta)\), recognizing that \(\theta\) is implicitly updated during training.
It comprises four hidden layers, each with 256 nodes, utilizing the Softplus activation with a high $\beta$ in intermediate layers to ensure smooth gradient updates while preserving non-linearity. The final output layer remains linear to maintain the integrity of the SDF values. 
Finally, inspired by Mip-NeRF 360 \cite{barron2022mip} \& iSDF\cite{ortiz2022isdf}, we also apply positional embeddings(PE) to the input coordinates $\xi$ and concatenate the embedding at an intermediate layer (the third layer) to better capture high-frequency detail.

\subsubsection{Loss Function}
\label{subsubsec:loss} 
The loss function \( \mathcal{L}(h_\text{sdf}) \) of our framework consists of three key components: SDF loss, gradient loss, and Eikonal loss.
The SDF loss supervises the network by enforcing consistency between the predicted SDF and the estimated ground truth. It consists of two terms: the free space loss and the near-surface loss. The free space loss, \( L_{\text{free\_space}} \), applies to points in free space and is defined as:
\begin{equation}\label{sdf_free_space}
L_{\text{free\_space}}(h_\text{sdf}(\xi), b) = \max(0, e^{-\beta h_\text{sdf}(\xi)} - 1, h_\text{sdf}(\xi) - b).
\end{equation}
Here, \( b \) represents the estimated ground truth SDF value. 
Since computing exact signed distances for all surface points is computationally expensive, we employ the batch distance method \cite{ortiz2022isdf} to approximate the ground-truth SDF efficiently while balancing performance and computational overhead.

Near the surface, the loss should be more stringent. The near-surface loss, \( L_{\text{near\_surf}} \), directly supervises the SDF prediction within these confines and is formulated using a huber loss\cite{huber1992robust}:
\begin{equation}\label{sdf_near_surface_huber}
\begin{aligned}
&L_{\text{near\_surf}}(h_\text{sdf}(\xi), b, \delta) =\\ 
&\begin{cases} 
\frac{1}{2}(h_\text{sdf}(\xi) - b)^2, & |h_\text{sdf}(\xi) - b| \leq \delta, \\
\delta (|h_\text{sdf}(\xi) - b| - \frac{1}{2}\delta), & |h_\text{sdf}(\xi) - b| > \delta.
\end{cases}
\end{aligned}
\end{equation}

Previous works like iSDF\cite{ortiz2022isdf} primarily use L1 loss, but our experiments demonstrate that huber loss mitigates the impact of noisy depth values, improving the quality of neural SDF reconstruction. A qualitative analysis comparing huber and L1 loss under real-world noisy depth settings is presented in section \ref{subsec:effect_depth_noise}.
The combined SDF loss is given by:
\begin{equation}
\begin{aligned}
&L_{\text{sdf}}(h_\text{sdf}(\xi), b) =\\
&\begin{cases} 
\lambda_{\text{surf}} L_{\text{near\_surf}}(h_\text{sdf}(\xi), b, \delta), & |D[u, v] - s| < \epsilon, \\
L_{\text{free\_space}}, & \text{otherwise}.
\end{cases}
\end{aligned}
\end{equation}
The second component, the gradient loss, refines the SDF prediction 
by supervising its gradient and ensure smooth, accurate gradient.
To enforce gradient consistency, the gradient loss penalizes the cosine distance between the predicted gradient and the approximated gradient \( \mathbf{g} \):
\begin{equation}
L_{\text{grad}}(h_\text{sdf}(\xi), \mathbf{g}) = 1 - \frac{\nabla_{\xi} h_\text{sdf}(\xi) \cdot \mathbf{g}}{\|\nabla_{\xi} h_\text{sdf}(\xi)\| \|\mathbf{g}\|}.
\end{equation}
The third component, the Eikonal loss, ensures that the learned model satisfies the Eikonal equation, enforcing a valid distance field:
\begin{equation}\label{Eikonal}
\begin{aligned}
&L_{\text{eik}}(h_\text{sdf}(\xi), \mathbf{g}) =\\
&\begin{cases}
0, & \|\nabla_{\xi} h_\text{sdf}(\xi)\| - 1 < \alpha', \\
\| \|\nabla_{\xi} h_\text{sdf}(\xi)\| - 1 \|, & \text{otherwise}.
\end{cases}
\end{aligned}
\end{equation}

where \( \alpha' \) is a threshold determining when the regularization is applied. The total loss function \( \mathcal{L}(h_\text{sdf}) \) is a weighted sum of all three terms:
\begin{equation}
\mathcal{L}(h_\text{sdf}) = L_{\text{sdf}} + \lambda_{\text{grad}} L_{\text{grad}} + \lambda_{\text{eik}} L_{\text{eik}}.
\end{equation}
where \( \lambda_{\text{grad}} \) and \( \lambda_{\text{eik}} \) control the relative influence of gradient and Eikonal constraints. 

\vspace{2pt}

\subsubsection{Continuity \& Differentiability of Neural SDF}

As noted in section~\ref{subsubsec:neural_arch}, our neural SDF \(h_\text{sdf}(\xi)\) first embeds the input \(\xi \in \mathbb{R}^3\) into a higher-dimensional space via PE, which employs smooth (infinitely differentiable) \(\sin\) and \(\cos\) functions \cite{barron2022mip}. We then feed \(\gamma(\xi)\) into an MLP with \(H\) hidden layers. 
At each layer \(i\), an affine map \(\mathbf{z}_i = \mathbf{W}^{(i)}\mathbf{q}_{i-1} + \mathbf{c}^{(i)}\) 
is followed by a \(\operatorname{Softplus}\) activation \(\mathbf{q}_i = \operatorname{Softplus}\bigl(\mathbf{z}_i;\beta\bigr)\), where \(\operatorname{Softplus}(z;\beta) = \frac{1}{\beta}\ln\bigl(1 + e^{\beta z}\bigr)\) \cite{glorot2011deep}. We set the initial layer to \(\mathbf{q}_0 = \gamma(\xi)\). 
The final output layer remains linear to preserve the signed-distance property:
\begin{equation}\label{eq:f}
    h_\text{sdf}(\xi) 
    \;=\;
    \mathbf{w}^{\top}\,\mathbf{q}_H 
    \;+\;
    \mathbf{c}.
\end{equation}

Because \(h_\text{sdf}(\xi)\) is composed of continuous functions (positional embedding, affine transformations, \(\operatorname{Softplus}\)), it is continuous on \(\mathbb{R}^3\). Moreover, each component is differentiable, 
so by the chain rule, the gradient \(\nabla h_\text{sdf}(\xi)\) exists and remains continuous everywhere \cite{hornik1989multilayer, park2019deepsdf, tancik2020fourier}. Concretely,
\begin{equation}\label{eq:grad}
\nabla h_\text{sdf}(\xi) 
\;=\;
\frac{\partial h_\text{sdf}}{\partial \mathbf{q}_H}\,
\frac{\partial \mathbf{q}_H}{\partial \mathbf{z}_H}
\,\cdots\,
\frac{\partial \mathbf{z}_1}{\partial \mathbf{q}_0}\,
\frac{\partial \mathbf{q}_0}{\partial \xi},
\end{equation}
where, for instance, \(\frac{d}{dz}\operatorname{Softplus}(z;\beta) = \frac{\beta\,e^{\beta z}}{1 + e^{\beta z}}\). 
Hence, \(h_\text{sdf}(\xi)\) and its gradient \(\nabla h_\text{sdf}(\xi)\) are both continuous and differentiable, satisfy the requirements needed for standard CBF formulations.

\subsection{RNBF Pipeline} 

Unlike existing work that twists standard CBF-QP formulations (Eq.~\eqref{eq:cbf-qp affine},~\eqref{eq:cbf-qp affine_1}) to accommodate noises and instability in the learned SDFs, we are able to combine SDF outputs from the vision pipeline smoothly with existing reactive safe controller, like CBF-QP, since our SDF learning process has explicitly considered sensor noises. Assuming robot state $x$ contains robot position, i.e. $x = [\xi; \xi']$ and $\xi'$ are robot state elements that exclude 3D position, the CBF-QP in~\eqref{eq:cbf-qp affine},~\eqref{eq:cbf-qp affine_1} can be rewritten as:
\begin{gather}
u_{\text{cbf}} = \argmin_{{u} \in \mathbb{R}^m}\frac{1}{2}||u-u_{\text{nom}}||_2^2\\
\nabla_\xi h_\text{sdf}(\xi)\dot{\xi} \geq -\alpha (h_\text{sdf}(\xi))\label{eq:cbf-qp vision}.
\end{gather}
As stated in section~\ref{sec: assumptions}, we reasonably assume that $h_\text{sdf}(\xi_\text{rob}) = h_\text{sdf}(\xi_\text{cam}) = h_\text{sdf}(\xi)$. It is noteworthy that while given any control affine robot dynamics, $\dot{x}$ is guaranteed to be actuated, i.e. $\dot{x}$ explicitly depends on control input $u$. However, such guarantee does not necessarily hold for $\dot{\xi}$, in which case an appropriate robot model must be chosen. Our resulted outputs from the vision pipeline can be adopted by any robot safe navigation algorithms that plan collision-free trajectories given real-time distance and the surface normal (or gradient) to the nearest obstacle. 

The overall RNBF-Control pipeline can be found in \autoref{fig:sdf_pipelinef}. where the SDF generation algorithms are running at a slower rate of 5-15 Hz and the reactive safe controller concurrently at a higher frequency to ensure robot safety. 
The vision pipeline processes RGB-D camera inputs as discussed in section \ref{subsec:SDF_Generation}. As the robot navigates, the vision pipeline continuously updates the environment’s SDF, while the control pipeline receives the current SDF \(h_\text{sdf}(\xi)\), and its gradient \(\nabla h_\text{sdf}(\xi)\) by querying neural SDF at the current camera position \(\xi\).
These inputs are fed into the CBF-QP algorithm, which calculates the necessary control commands. These commands adjust the robot's position within the environment, facilitating safe navigation towards the target location. 


\section{Experiments}
\subsection{Experiment Setup}
\label{subsec:effect_depth_noise} 

\begin{figure}[t]
  \centering
  \subfloat[]%
    {\includegraphics[width=0.48\linewidth]{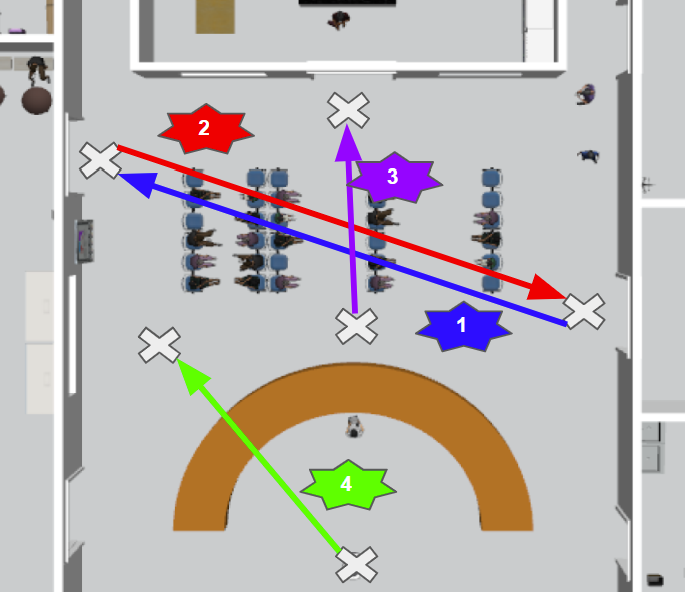}\label{fig:setup-a}}\hfill
  \subfloat[]%
    {\includegraphics[width=0.48\linewidth]{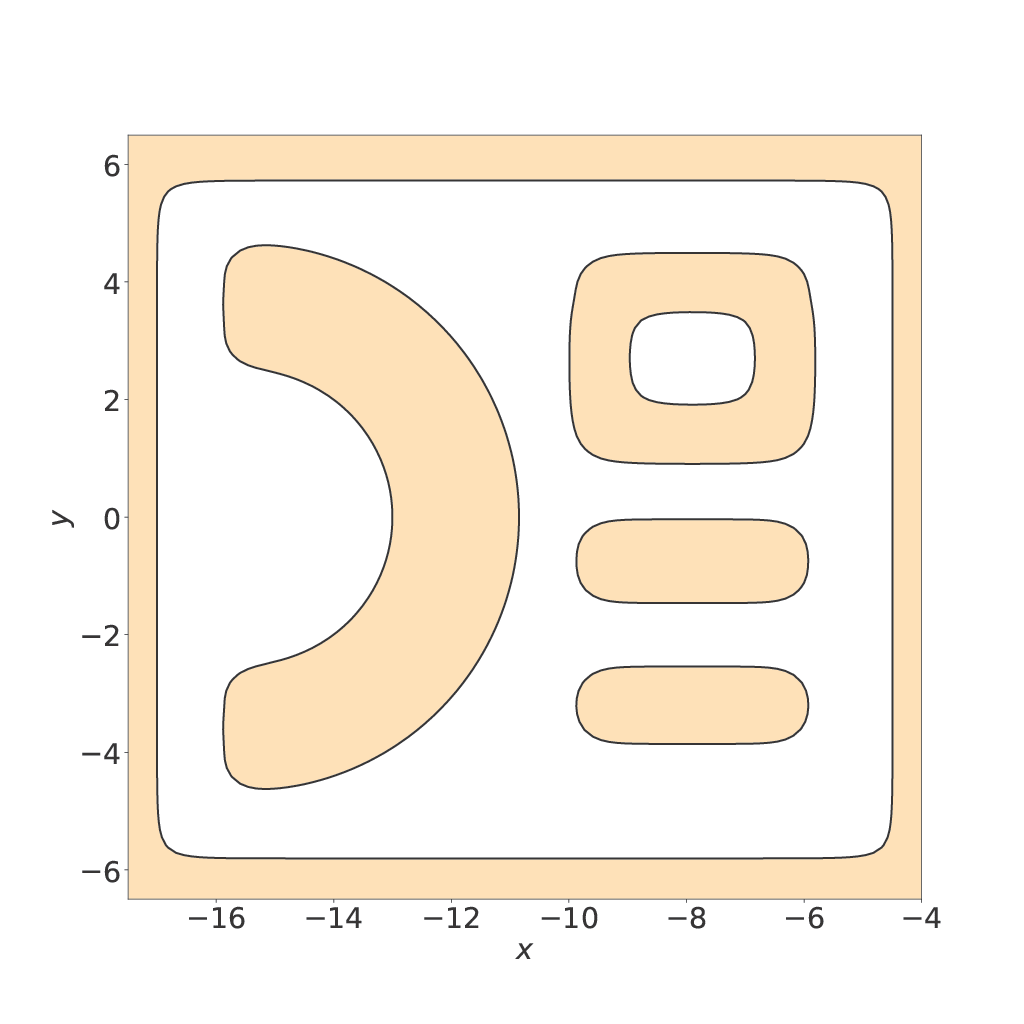}\label{fig:setup-b}}
  \caption{
  (a) Simulation environment setup with four different scenarios.
  (b) Ground-truth SDF used by Parametric CBF-QP. Note, RNBF-CBF-QP builds its SDF online and does not use the ground-truth SDF.}
  \label{fig:setup}
\end{figure}

Our proposed approach is evaluated on a Fetch robot running at 20 Hz in simulation and robot hardware. For simulations, we employ a realistic hospital environment in Gazebo, as shown in \autoref{fig:setup-a}, and define four scenarios with different initial and target combinations. For real-world validation, we equip the Fetch robot with an Intel RealSense D435i RGB-D camera (\autoref{fig:fetch}) and utilize a motion capture system to acquire real-time camera poses. All experiments are run on an Intel Core i7-12700K CPU with a GeForce RTX 3090 Ti GPU.

\begin{figure}
    \centering
    \includegraphics[width=0.40\linewidth]{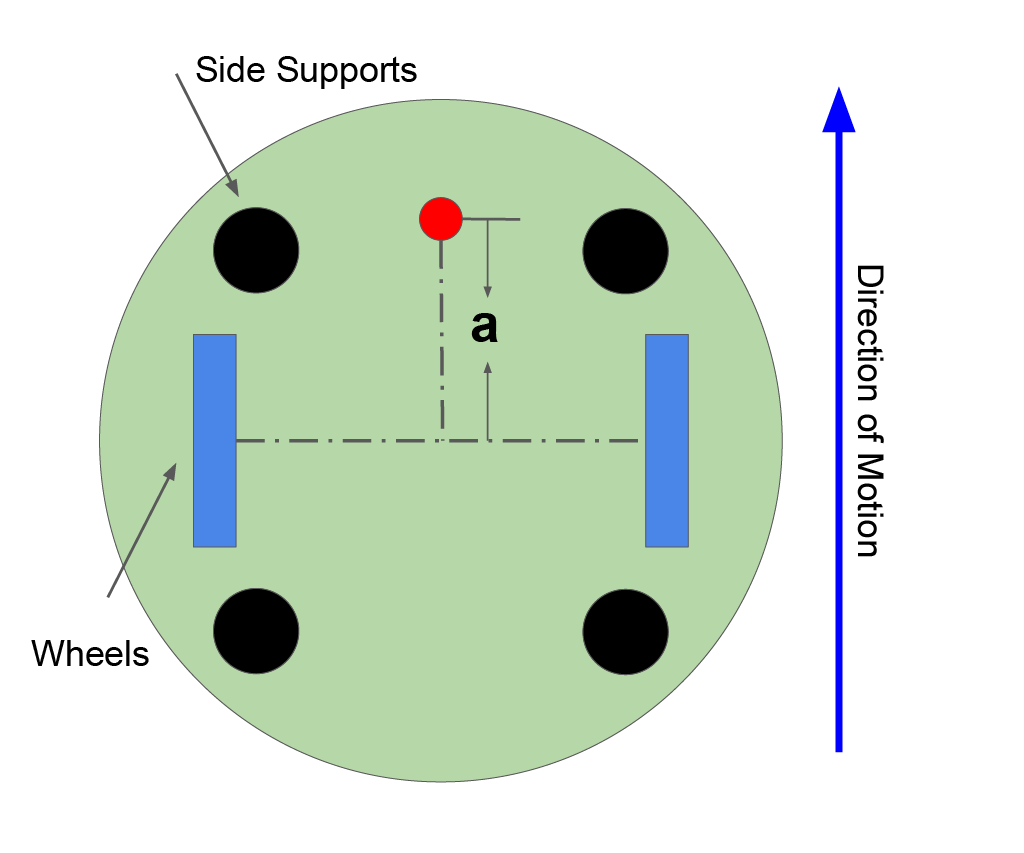}
    \caption{Modified point of interest on Fetch robot.}
    \label{fig:shift point}
    \vspace{-5mm}
\end{figure}
\begin{figure*}[t]
\centering
\includegraphics[width=0.95\linewidth]{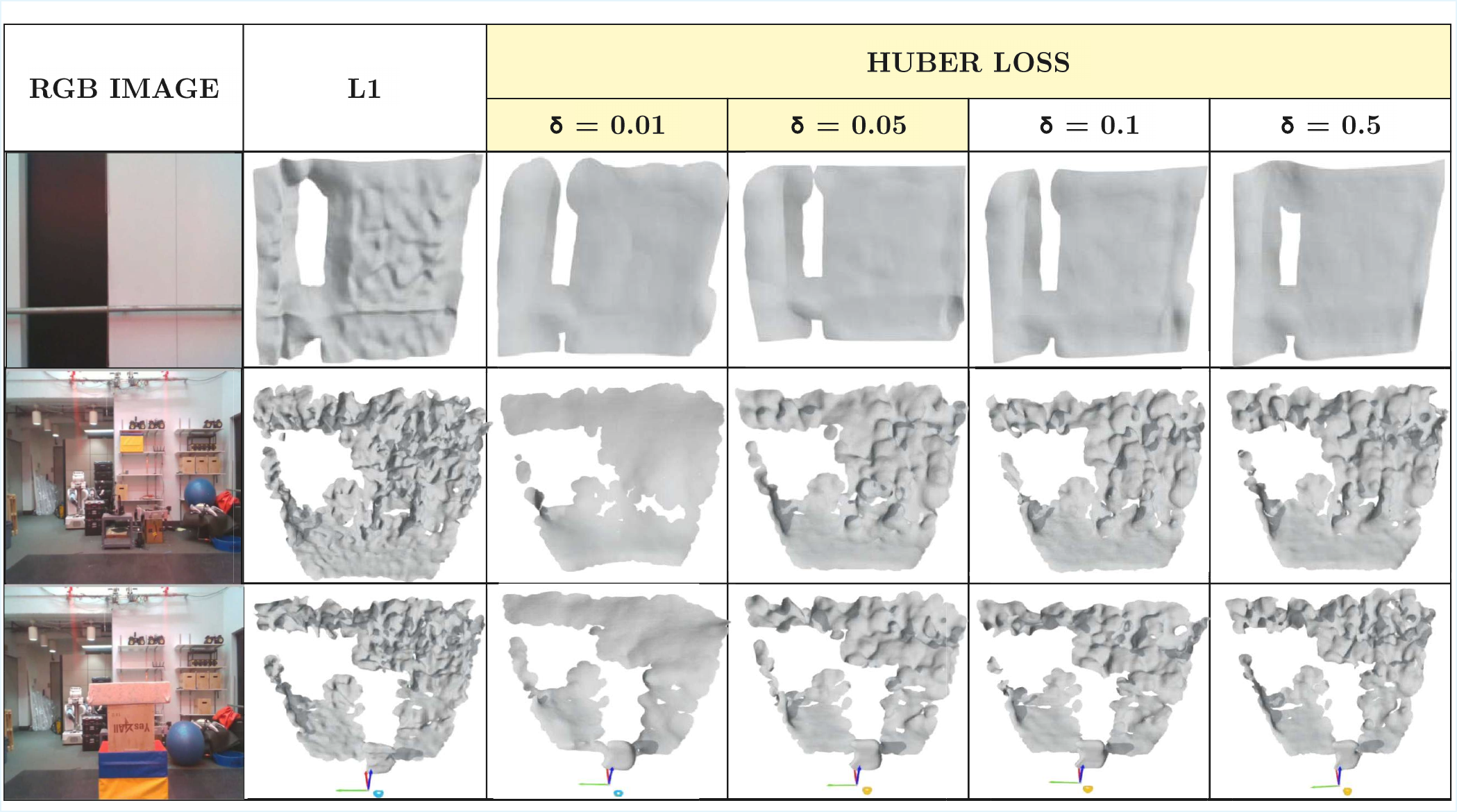}
\caption{Comparison of SDF reconstructions using L1 loss \& huber loss functions on depth data from an Intel RealSense D435i sensor. The leftmost column shows the RGB images of the scene. The second column presents reconstructions using the L1 loss. The remaining columns show reconstructions using the huber loss with varying $\delta$ values. Each row corresponds to a scene with a different camera-to-surface distance: in the first row, the wall is approximately 2 meters from the camera; in the second row, the racks and back wall are approximately 7 meters away; and in the third row, the closest obstacle is approximately 3 meters away.} \label{fig:Depth_Noise}
\vspace{-5mm}
\end{figure*}

\subsection{Robot Dynamics}
Differential drive robots like Fetch is often modeled as a standard unicycle \eqref{eq:standard unicycle}. 
 \begin{equation}
\label{eq:standard unicycle}
    \dot{x}=\begin{bmatrix}
        \dot{p}_x\\\dot{p}_y\\\dot{\theta}
    \end{bmatrix}=
    \begin{bmatrix}
        \cos{\theta} & 0\\ \sin{\theta} & 0\\ 0 & 1
    \end{bmatrix}
    \begin{bmatrix}
        v \\ \omega
    \end{bmatrix}
\end{equation}

 However, in our experiment, we choose to instead model Fetch using the shifted unicycle model by selecting a point of interest $a>0$ m ahead of the wheel axis (see \autoref{fig:shift point}). The unicycle model with the shifted coordinate system becomes the following. 
\begin{equation}
\label{eq:shifted unicycle}
    \dot{x}=\begin{bmatrix}
        \dot{p}_x\\\dot{p}_y\\\dot{\theta}
    \end{bmatrix}=
    \begin{bmatrix}
        \cos{\theta} & -a\sin{\theta}\\ \sin{\theta} & a\cos{\theta}\\ 0 & 1
    \end{bmatrix}
    \begin{bmatrix}
        v \\ \omega
    \end{bmatrix}
\end{equation}
Note that for differential drive robot, $x = [p_x, p_y, \theta]^\top$, $\xi = [p_x, p_y]^\top$, and $u = [v, \omega]^\top$. Position in the z-axis is neglected for 2D navigation tasks. Our robot model decision is made due to the fact that $\dot{\xi}$ using the standard unicycle model depends on control input $v$ while using the shifted model depends $\dot{\xi}$ on both $v$ and $\omega$. The CBF constraints for the shifted unicycle system can be written as
\begin{equation}
\begin{aligned}
\frac{\partial h(p_x,p_y)}{\partial p_x}\dot{p}_x + \frac{\partial h(p_x,p_y)}{\partial y}\dot{p}_y\geq
-\alpha (h(p_x,p_y)).
\end{aligned}
\end{equation}

\textbf{Nominal Controller:} Our nominal controllers are designed to approximately follow linear trajectories $u_\text{nom}^l$ pointing from the robot's position $\xi$ to the target $\xi^*$ with a speed of 0.5 $m/s$, where $\psi$ is the angle difference between the current robot pose $\theta$ and the desired robot pose estimated using the orientation of $u_\text{nom}^l$, and $\Delta t$ is the update timestep of the nominal controller. The approximation is not exact due to the non-holonomic property of differential drive systems. 

\begin{equation}
\begin{aligned}
u_\text{nom}^l = 0.5\frac{\xi-\xi^*}{||\xi-\xi^*||_2} \quad
v_\text{nom} = ||u_\text{nom}^l||_2 \quad \omega_\text{nom} = \frac{\psi}{\Delta t}
\end{aligned}\label{eq:dubin nominal system}
\end{equation}

\subsection{Effect of Depth Noise on SDF}
\label{subsec:effect_depth_noise} 

When deploying our pipeline on a real robot, we use a low-cost Intel RealSense D435i sensor, whose depth estimates are notably noisy. Such noises introduce spurious artifacts that would degrade the SDFs and their gradients from the neural network if left attended. Such artifacts will disqualify the SDFs from being approximate barrier functions. To mitigate this issue, we adopt a huber loss for \(L_{\text{near\_surf}} \) (\ref{sdf_near_surface_huber}), which reduces the impact of outliers while still enforcing tight supervision near the surface. \autoref{fig:Depth_Noise} compares SDF reconstructions using RealSense depth data. 
In this work, sensor noise refers specifically to the depth noise inherent in affordable RGB-D sensors.
The L1 loss based model amplifies sensor noise, leading to rough surfaces and unstable gradients, whereas the huber-based model produces smoother geometry and more stable gradients. The choice of \( \delta \) significantly influences reconstruction quality: smaller values of \( \delta \) suppress noise more effectively but risk oversmoothing fine details, while larger values retain geometric features but may allow noise to propagate. In our real-world experiments, we find \( \delta = [0.01, 0.05] \) to offer a favorable balance, though it remains a tunable parameter depending on the quality of depth sensor. 
Although we do not claim perfect fidelity, the result is sufficiently accurate for our pipeline. Overall, the huber loss preserves the key property of a smoothly varying SDF, essential for real-time CBF-based control under imperfect depth measurements.

\subsection{Experiment Results}

\begin{figure}[!tbp]
    \centering
    \includegraphics[width=0.75\linewidth]{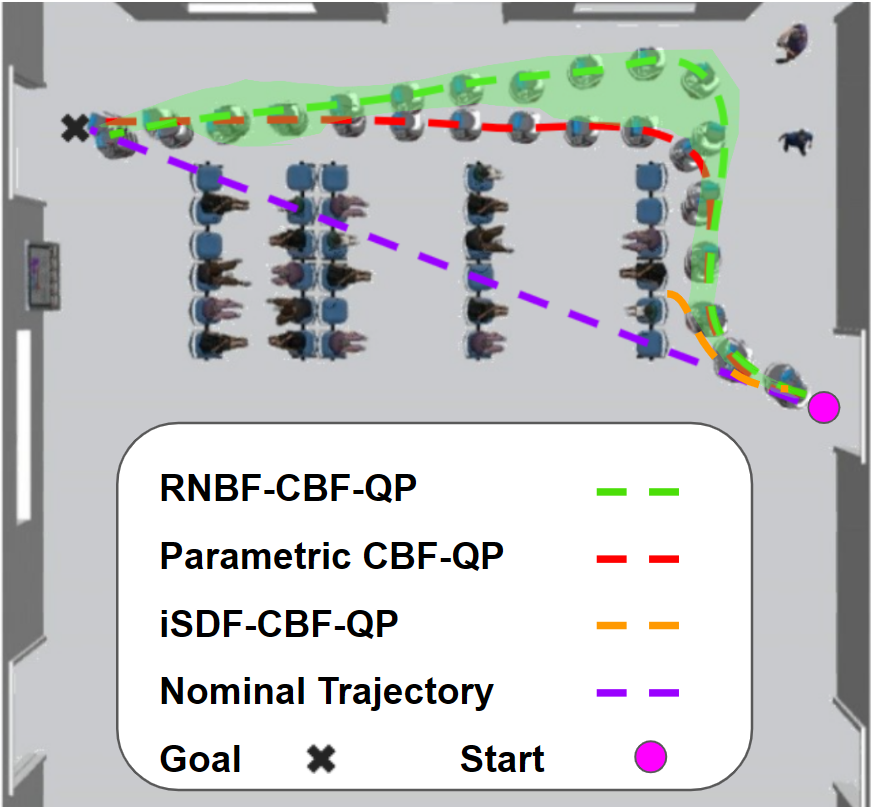}
    \caption{
    \textbf{Scenario 1 (simulation): Trajectory comparison.}
    Dashed curves represent trajectories, with mean values shown where applicable. For RNBF-CBF-QP, a shaded tube depicts the variance of the mean trajectory over 10 rollouts. Parametric CBF-QP, which assumes perfect obstacle knowledge, produces nearly identical paths across runs; only one representative trajectory is plotted for clarity. iSDF-CBF-QP is likewise shown with a single representative run since it collided in all trials. The nominal controller is deterministic and follows the shortest path from start to goal.}

    \label{fig:comp1}
    \vspace{-5mm}
\end{figure}


In this section, we present the experimental results that highlight the effectiveness of our \textit{RNBF-based safety filter}. While in this work we demonstrate its use with a CBF-QP controller \textit{(RNBF-CBF-QP)}, the RNBF-based safety filter pipeline can be readily adapted to any SDF-based controller or safety filter that consumes an SDF and its gradients. RNBF-CBF-QP is compared against two baselines: (i) \textit{Parametric CBF-QP}, which assumes full knowledge of all obstacle locations and geometries using the ground-truth SDF of the environment (\autoref{fig:setup-b}), and 
(ii) \textit{iSDF-CBF-QP}, which integrates the original iSDF reconstruction into a CBF-QP without floor segmentation or depth-denoising. We display in \autoref{fig:comp1} the trajectories generated by RNBF-CBF-QP, the parametric CBF-QP, iSDF-CBF-QP and the nominal controller for scenario 1.

\begin{table}[h]
    \centering
    \renewcommand{\arraystretch}{1.2} 
    \begin{tabular}{|l|c|c|c|}
        \hline
        \textbf{Methods} & \textbf{Safe \%} & \textbf{Reached \%} & \textbf{Duration (s)} \\
        \hline
         RNBF-CBF-QP & 100 & 100 & 36.7 \\
        \hline
         Parametric CBF-QP & 100 & 100 & 29.0 \\
        \hline
        iSDF-CBF-QP & 0 & 0 & - \\
        \hline
    \end{tabular}
    \caption{Performance comparison of RNBF-CBF-QP with Parametric CBF-QP \& iSDF-CBF-QP in Scenario 1 over 10 independent trials.}
    \label{tab:cbf_results}
\end{table}

\begin{figure}[h!]
    \centering
    \includegraphics[width=0.95\linewidth]{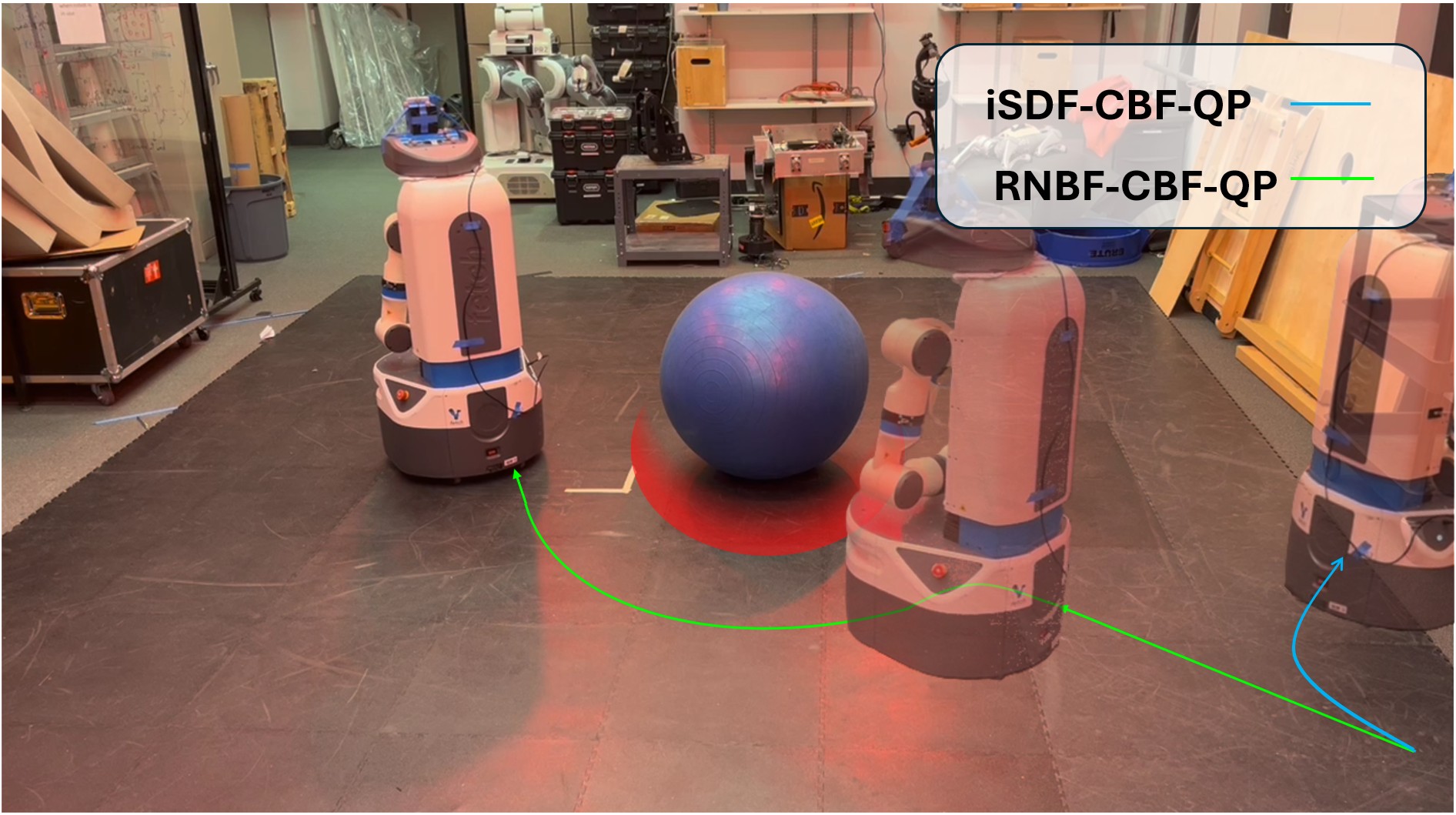}
    \caption{Trajectory comparison of RNBF-CBF-QP vs iSDF-CBF-QP for real-world deployment with static obstacle.}
    \label{fig:isdf_vs_rnbf}
    \vspace{-5mm}
\end{figure}

\autoref{tab:cbf_results} reports quantitative results for ten independent runs of Scenario 1 with RNBF-CBF-QP, Parametric CBF-QP, and iSDF-CBF-QP. Performance is measured by the average traversal time from start to goal (\emph{duration}, s), the proportion of collision-free trajectories (\emph{safe \%}), and the proportion of successful goal reaches (\emph{reached \%}). Both RNBF-CBF-QP and the Parametric CBF-QP achieved a 100\% collision-free rate and a 100\% success rate in reaching the destination. However, RNBF-CBF-QP requires slightly more time to reach the goal (36.7\,s vs.\ 29.0\,s) owing to its exclusive reliance on online RGB-D sensing to construct barrier functions, as opposed to the Parametric CBF-QP’s use of full prior knowledge of obstacle locations. 
In contrast, the iSDF-CBF-QP fails in all trials, with $0\%$ success rate in reaching the destination and repeated collisions with the obstacles. Two factors primarily drive this failure: 
(1) \textit{Floor misclassification}: without a segmentation module, large planar floor regions are incorrectly treated as obstacles, preventing the controller from finding a feasible path; this issue arises in both simulation and real-world deployments.
(2) \textit{Depth-noise artifacts}: sensor noise introduces spurious structures into the neural SDF (section \ref{subsec:effect_depth_noise}), yielding unstable gradients and erratic control commands; while less relevant in simulation due to noise-free gazebo depth sensing, it significantly impacts physical robot navigation (\autoref{fig:isdf_vs_rnbf}). RNBF-CBF-QP mitigates both issues through explicit floor segmentation and huber-loss based near-surface supervision, enabling stable gradients and reliable, collision-free navigation.

\begin{figure}[!tbp]
    \centering
    \includegraphics[width=0.95\linewidth]{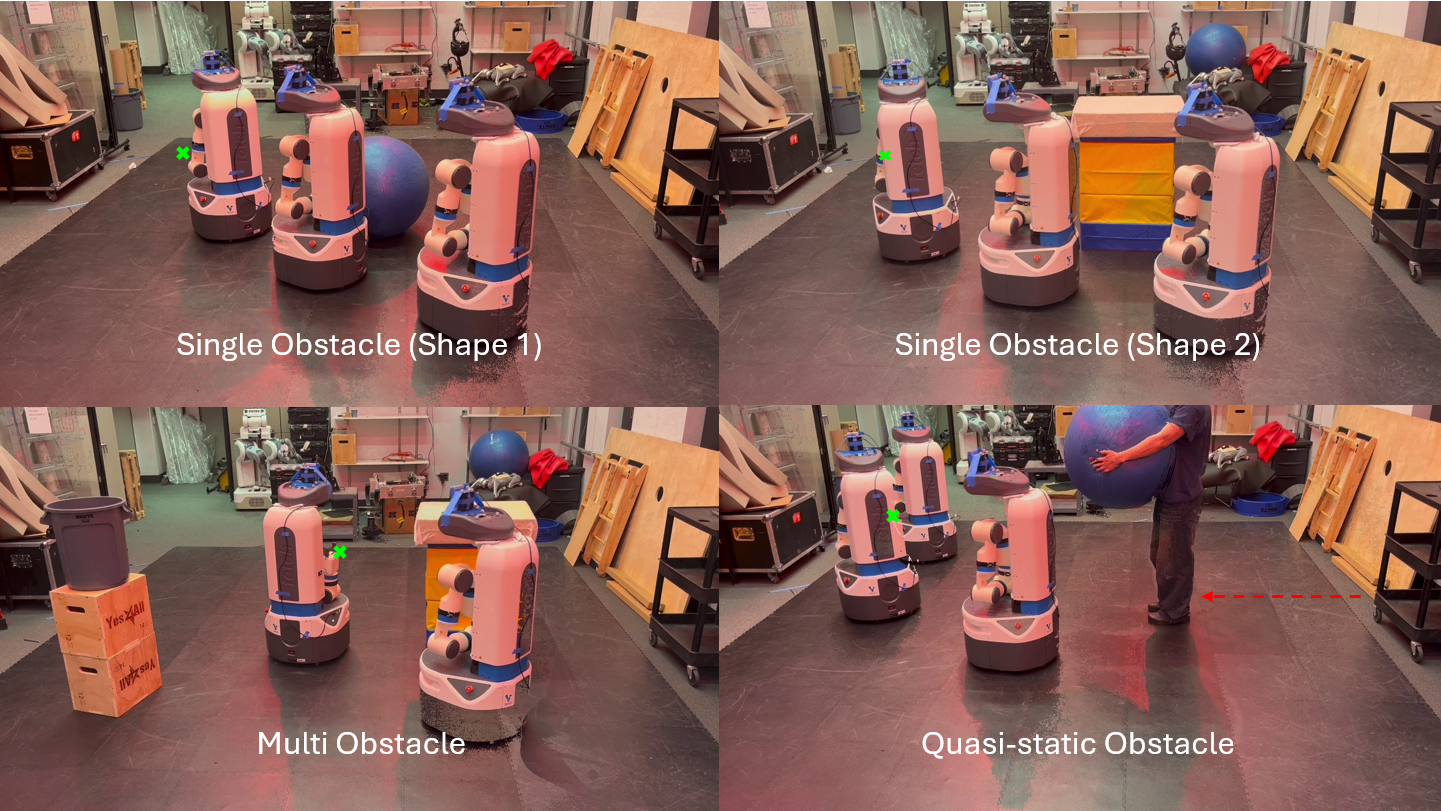}

  \caption{\textbf{Real-world experiments.} Shown are collision-free RNBF-CBF-QP trajectories overlaid on video frames for static (single obstacle - shape 1 \& 2, and multi obstacle) and quasi-static obstacle scenarios.}

    \label{fig:real_world_traj}
    \vspace{-5mm}
\end{figure}

Beyond gazebo simulations, the proposed pipeline is further validated in real-world experiments involving both static and quasi-static obstacles (\autoref{fig:fetch} \& \autoref{fig:real_world_traj}). 
For both types of obstacles, RNBF-CBF-QP reached the goal collision-free under partial observability and depth noise. In contrast, iSDF-CBF-QP exhibited the same failure modes as in simulation leading to collisions and failure to reach the goal as shown in \autoref{fig:isdf_vs_rnbf}.
We provide supplementary videos\footnote{
Videos and supplementary material are available at
\url{https://satyajeetburla.github.io/rnbf/}}, which reveal that our method is capable of accurately \textit{building and updating the barrier function on-the-fly}, thereby ensuring safety in novel environments.

\section{CONCLUSIONS}
The paper introduces a novel pipeline for real-time CBF construction in unknown environments using affordable sensors like RGB-D cameras. Our framework is robust to sensor noise while generating a smooth and differentiable neural SDF representation, thereby preserving the core assumptions required by standard CBFs. Experiments in both simulation and on a physical Fetch robot confirm that our approach enables safe, reactive navigation 
and can readily integrate with established SDF-based controllers. Moreover, the system’s modular design allows it to serve as a dedicated safety layer in broader robotic applications, given only onboard RGB-D sensing.

Although our method demonstrates robust performance in static and quasi-static environments, it faces limitations in fully dynamic settings. Currently, our RNBF pipeline lacks a “forgetting” mechanism, causing outdated obstacle data to persist in the reconstructed map and reducing adaptability to rapidly changing scenarios. As part of future work, we plan to incorporate real-time memory management to discard stale obstacle information, as well as integrate temporal reasoning and motion prediction to further enhance performance in dynamic environments.











\bibliographystyle{IEEEtran}
\bibliography{citations.bib}

\end{document}